\documentclass[conference]{IEEEtran}
\ifCLASSINFOpdf
\else
\fi
\begin{document}
\title{Visual Anomaly Detection for Images: A Survey}



\author{\IEEEauthorblockN{Jie Yang,
Ruijie Xu,
Zhiquan Qi, 
Yong Shi}
\IEEEauthorblockA{University of Chinese Academy of Sciences, Beijing 101408, China,
\\ Email: yangjie181@mails.ucas.ac.cn}
}


\maketitle

\begin{abstract}
Visual anomaly detection is an important and challenging problem in the field of machine learning and computer vision. 
This problem has attracted a considerable amount of attention in relevant research communities. 
Especially in recent years, the development of deep learning has sparked an increasing interest in the visual anomaly detection problem and brought a great variety of novel methods.
In this paper, we provide a comprehensive survey of the classical and deep learning-based approaches for visual anomaly detection in the literature. 
We group the relevant approaches in view of their underlying principles and discuss their assumptions, advantages, and disadvantages carefully.
We aim to help the researchers to understand the common principles of visual anomaly detection approaches and identify promising research directions in this field.
\end{abstract}

% \begin{keyword}
% Visual anomaly; visual anomaly detection; deep learning.
% \end{keyword}

% \fi
\IEEEpeerreviewmaketitle

\section{Introduction}
\label{main}
Anomaly detection is an important machine learning problem. Different from the assumption of static and closed system, that most existing machine learning methods are based on, it researches how machine learning models can deal with the unknown and uncertain information under the open and dynamic system environment. With the assumption of open environment, the learning systems developed for anomaly detection are usually expected to leverage the knowledge from the knows (normal data and patterns) to infer the unknowns (abnormal or novel patterns that different from the normal ones). Anomaly detection approaches usually extract, characterize and model the patterns with the available normal data, and then develop reasonable anomaly detectors to discover novel or abnormal patterns in the newly observed data.
When the target of anomaly detection is the image data, then comes the visual anomaly detection or image anomaly detection. 

The research of visual anomaly detection is important both in view of the theory and applications. On the one hand, as mentioned, since the research paradigm of visual anomaly detection is beyond the assumption that building machine learning models in a closed systems, it will drive the development of artificial intelligence in both theory and application. On the other hand, visual anomaly detection has a broad application prospect. For example, in the field of intelligent manufacturing, it can be applied to defects detection; in the field of medical image analysis, it can be used to detect the lesions in medical images; in the field of intelligent security, it can be used to detect the abnormal events in videos.

This paper will provide a comprehensive review of the literature about visual anomaly detection. We will group the relevant approaches in view of their basic ideas and review each of them carefully following their developmental logic. Generally, the categorization and review of the various visual anomaly detection approaches in the current literature will help relevant researchers to understand the development and further trend of visual anomaly detection, develop novel detection techniques and open up new research directions.

\section{Overview of visual anomaly detection}
From the perspective of whether the supervised information (whether there are abnormal samples or abnormal patterns) is available, Visual anomaly detection can be divided into two different research themes: supervised and unsupervised visual anomaly detection. In this paper, we will mainly review the approaches that address the unsupervised visual anomaly detection problem. On the one hand, unsupervised visual anomaly detection drives the proposed approaches to develop computer vision models that can capture the detection ability as our human vision. On the other hand, the unsupervised situation is more common for most practical application scenarios. Because in many application scenes, abnormal image samples are very rare and difficult to collect, there is no effective supervision information to use. Moreover, abnormal samples or abnormal patterns are usually variable in shape, color and size, and they do not have stable statistical laws. All these will make it difficult for the supervised learning model to capture enough statistical information or salient features about abnormal image patterns.

Besides, in view of different visual detection granularity, visual anomaly detection can be grouped into two categories: image-level and pixel-level visual anomaly detection. Among them, image-level detection usually only focuses on the question of whether the whole image is normal or abnormal, while pixel-level anomaly detection further requires detect or locate the abnormal regions in the image.

In addition, according to the historical development of visual anomaly detection researches, the literature for unsupervised image (including image-level and pixel-level) anomaly detection can be roughly divided into two stages: before deep learning and after deep learning. 
Before the deep learning is put forward, the research for visual anomaly detection focus on developing anomaly detection strategies or mechanisms.
The main research problem is: After obtaining the shallow features of images by hand, such as the gray value, SIFT~\cite{lowe1999object} and HOG~\cite{dalal2005histograms}, it attempts to develop different the detection mechanisms based on statistics or traditional machine learning methods, such as density estimation, one-class classification and image reconstruction. It estimates the distribution model of normal images or the images features first. Then, if the images or the features do not meet the corresponding distribution model, they will be recognized as the anomalies.
After the development of deep learning techniques, especially after that the deep convolution neural networks have been made great success both in low-level and high-level computer vision tasks~\cite{img-restoration,aaai-jie,edge_hed,electronics2021likai,vgg19,resnet,semantic_seg}, the relevant researchers gradually shift their attentions to the question of how to combine the powerful representation capability of the deep convolution network with the problem of visual anomaly detection, and devote themselves to develop the end-to-end detection approaches.

In the next sections, we will follow the development of the research of unsupervised visual anomaly detection, group the relevant approaches into categories in view of the different detection mechanisms behind them and review the existing researches on image-level and pixel-level anomaly detection respectively.

\section{Image-level visual anomaly detection}
Based on the different detection mechanisms, unsupervised image-level anomaly detection methods can be roughly divided into four groups: density estimation, one-class classification, image reconstruction and self-supervised classification.

\subsection{Density estimation}
The density estimation method estimates the probability distribution model of the normal images or features firstly, then detects and identifies whether the newly observed image is abnormal or normal by testing against the established distribution. Methods based on density estimation usually assume that if the testing image or image feature does not meet the probability distribution model that is estimated with the normal image samples, it will be classified as anomaly. Concretely, density estimation methods first model the probability distribution of the normal images or features, then judge whether the test image is normal or abnormal by estimating the likelihood probability or score of the test image against the established distribution. 

Typical density methods include parameter distribution estimation, such as Gaussian model and Gaussian mixture model~\cite{2006-pr-book, 2009-one-class-survey}, non-parametric estimation methods, such as nearest neighbor and kernel density estimation method~\cite{2009-one-class-survey}. However, estimating a reasonable probability density requires a large number of training samples. And when the feature dimension of the samples is very large (such as image data), this problem of the number of training samples becomes particularly prominent. In addition, the scalability of these classic models is poor. The deep generative models~\cite{VAE, dinh2016density, akoury2017spatial, kingma2018glow} are capable of establishing the probability distribution of the high-dimension data, such as images. But they still need a large number of image samples for the related task. At the same time, deep generative models are not robust enough, and their performances for anomaly detection are not stable.
Through a large number of experiments, the authors in~\cite{nalisnick2018deep} found that the current popular deep generative models, such as variational automatic encoder (VAE)~\cite{VAE} or flow model (Glow)~\cite{kingma2018glow}, are not even competent for simple image anomaly detection tasks.

\subsection{One-class classification}
As the name suggests, one-class classification is to classify a single class, concretely which attempts to construct a decision boundary of the target class (normal images) in the feature space. Classic approaches are one-class support vector machines (OCSVM)~\cite{ocsvm} and support vector data description (SVDD)~\cite{svdd}. One-class methods do not need to estimate the concrete probability of each sample point of the image distribution, so they do not require a large number of training samples. But they still suffer from the problems of dimension disaster and scalability.

After the deep learning techniques are put forward, especially the deep convolution neural network, the relevant researchers gradually shift their attention on how to combine the deep convolution network with classic one-class classification approaches.
Perera et al.~\cite{occ-tf} propose an one-class classification method based on transfer learning, which fine-tunes the pre-trained convolution network to extract discriminative image features and then takes the nearest neighbor classification method to construct the one-class classifier. Perera et al.~\cite{occ-tf-m} further extend it to the problem of detecting anomalies for multiple known classes instead of one class. Since the approaches are not end-to-end deep classification models, their performance may be sub-optimal~\cite{dis-gan}.
Afterwards, Ruff et al.~\cite{dsvdd} propose an end-to-end deep support vector description model and use it for anomaly detection. Oza et al.~\cite{occnn} propose an end-to-end one-class classification neural network (OCCNN).

%In addition, some methods to improve the ability of deep classification network to deal with uncertain or out of distribution samples, although not specially proposed for image anomaly detection and recognition, can also be used for image anomaly detection, and can achieve good results~\cite{hendrycks2016baseline,lee2017training}.

\subsection{Image reconstruction}
Image reconstruction approaches map the image to a low-dimensional vector representation (latent space), and then try to find an inverse mapping or reconstruction for the original image. The reconstruction-based approaches for anomaly detection assume that the reconstruction errors of the normal images are small, while that of abnormal images are larger. The widely used reconstruction model is autoencoder. 
Autoencoder is proposed by Hinton~\cite{1989hinton} and the basic idea behind it is redundancy compression and non-redundancy separation.
Autoencoder is a neural network that has a narrow middle hidden layer. It tries to compress the input data through the hidden layer and then regenerates the input. Because the hidden layer is very narrow, the network may compress the redundant information in the input data while retain and distinguish the non-redundant information.
The model is also used to simulate the information processing of the hippocampus in our brains~\cite{1995ae-anomaly}. 

Japkowicz et al.~\cite{1995ae-anomaly} first introduce the autoencoder into the field of anomaly detection. Its working mechanism is that redundant information in normal data may not be redundant information in abnormal data and vice versa. Along with the development of deep learning, Sakurada et al.~\cite{sakurada2014anomaly} first use the deep autoencoder for anomaly detection of high-dimensional data. In addition to the reconstruction error, the reconstruction probability estimated by Monte Carlo sampling for scoring anomalies is proposed in ~\cite{an2015variational}.

At the same time, some scholars propose to leverage the distribution in the latent space and reconstruction errors of the deep autoencoder simultaneously, which can further improve the anomaly detection performance of the reconstruction-based models.
In~\cite{zong2018deep-gmm}, the author propose a Gaussian mixture model to estimate the distribution of the latent code of autoencoders.
Besides, the probability distribution of the latent code is modeled by an auto-regressive neural network in~\cite{ae-lsa}.
Moreover, in~\cite{ae-mm}, the authors introduce memory units for the latent code of the autoencoder, where a few feature memory units are exploited to represent the latent distribution.

In addition, some methods suggest to increase the difficulties for the image reconstruction. Typically, some transformations for the input image are firstly performed, such as color removal or geometric transformations. Then the autoencoder is trained to reconstruct the original input image with the incomplete input image on which the transformation is made~\cite{fei2020attribute,salehi2020puzzle}. This method can effectively increase the reconstruction difficulties for abnormal images, so their reconstruction errors are usually large. As a result, it can effectively increase the disparities between normal images and abnormal images, thus improving the performance of anomaly detection.

Schleglv et al.~\cite{anogan} is the first to adopt generative adversarial network (GAN) for visual anomaly detection. The GAN model is trained by normal image beforehand, and then the anomaly can be detected by calculating the difference between the test image and the normal image that is the closest to the test image. The closest normal image is found by performing an iterative optimization process. Firstly, we try to find the closest latent code for the test image in the GAN's latent space with the gradient descent strategy, and then use the GAN model to generate the corresponding normal image with the obtained latent code. Since this model needs to perform an iterative search process, its efficiency is usually unsatisfactory in practice.

Some approaches combine image reconstruction and adversarial training to further improve the performance and efficiency of visual anomaly detection. Sabokrou et al.~\cite{adv-one-class} is the first to leverage autoencoder and adversarial training simultaneously for image anomaly detection. Their model include a generator which is a convolutional autoencoder and a discriminator. They optimize the model both with the mean square error loss and the adversarial loss. In the testing stage, the reconstruction error and the output probability of discriminator are used as the indications for detection. Ak{\c{c}}ay et al.\cite{akccay2019skip} further introduce an encoder where the difference between the latent code of reconstructed image and the ones of the input image is also used as an indicator for the anomaly.
Zenati et al.~\cite{zenati2018efficient} adopt BiGAN~\cite{donahue2016adversarial} for image anomaly detection.
In~\cite{ocgan}, the authors propose to constrain or regularize the feature distribution of the latent code in the joint model of autoencoder and adversarial network. The approach further improves the performance of visual anomaly detection.
In order to overcome the disadvantage of lacking robustness when using discriminator's output of an GAN model as the anomaly score, a new training strategy is proposed in~\cite{zaheer2020old}. The training strategy can effectively improve the stability of the probability output of the discriminator.

\subsection{Self-supervised classification}
Visual anomaly detection by self-supervised classification mainly owes to the powerful visual feature representation capability of self-supervised learning. Self-supervised learning usually leverages certain auxiliary task (pretext) and attempts to mine the available supervision information from large-scale unsupervised data. It takes full use of the self-constructed supervision information to learn the visual representation typically by a deep convolutional neural network, and then the representations can be transferred to many downstream tasks, such as image classification, object detection and anomaly detection~\cite{rotate,2020Self-survey,2019-self-pretrain}.

The idea behind self-supervised classification for anomaly detection is that through self-supervised training, models can learn unique and more significant characteristics and features of normal samples. The representations learned for the target objects not only reflect their color, texture and other low-level features, but also reflect the high-level features such as the location, shape, position and direction.
By learning these features only from normal samples, the model then can effectively discern abnormal samples without such characteristics.
Golan et al.~\cite{rotate} proposes RotateNet, which makes a neural model learn to distinguish the different geometric transformations (rotations) that performed on normal images. This approach is the first to detect image anomalies by constructing a self-supervised classification model.

%, which should be able to locate the significant object in the image and identify the direction and type of the object

Ali et al.~\cite{2020Self-survey} compared various self-supervised methods and found that the self-supervised classification method, i.e. RotateNet~\cite{rotate}, shows the state-of-the-art performance for anomaly detection. In order to predict the rotation of the image, RotateNet must learn about the shape, location and direction of the target object in the image. Obviously, only learning the low-level features such as color and texture is not enough to predict the rotation of the image. Therefore, unlike other self-supervised representation learning methods (such as image reconstruction), RotateNet learns about both of the low-level and high-level features, thus performing better on the visual anomaly detection task. However, for those symmetrical objects, RotateNet may fail to predict the rotation and capture effective representations for anomaly detection.
Dan et al.~\cite{2019-self-pretrain} further propose to increase the difficulty of the task for self-supervised classification, i.e. not only distinguishing the rotation of the image but also classifying the translation of the image.
Tack et al.~\cite{tack2020csi} introduce contrastive learning (a kind of self-supervised classification method) for image anomaly detection and show that it performs better when compared with other self-supervised classification methods.
Sehwag et al.~\cite{sehwag2021ssd} further improve and extend the contrastive learning for anomaly detection and achieve the state-of-art image-level anomaly detection performance.

\section{Pixel-level visual anomaly detection}
In terms of different anomaly detection mechanisms, approaches for unsupervised pixel-level anomaly detection can be roughly divided into two categories: image reconstruction and feature modeling. It should be pointed out that all image-level anomaly detection methods generally can be used for pixel-level anomaly detection, because we can segment the whole image into multiple pieces, and then perform image-level anomaly detectionon the image patches. However, image-level anomaly detection algorithms focus on the semantic information of the whole image. But the semantic information in each image patch may be weak, so the applicability of image-level detection may be questionable. In the next sections, the methods specifically developed for pixel-level unsupervised anomaly detection will be reviewed in detail.

\subsection{Image reconstruction}
A typical method of image reconstruction is to compress and reconstruct the input image with the deep convolution autoencoder~\cite{mvtec,ae-ssim-l2,ae-texture}. 
It first learn to reconstruction of the normal images. Then, potential anomalies are detected by evaluating the pixel difference between the input image and the reconstructed image, where pixel-level~$l_2$-distance~\cite{l2}and image structure similarity measure (SSIM)~\cite{SSIM} can be used to measure the disparities before and after the reconstruction.
Generally, reconstruction-based methods assume that the convolutional autoencoder trained only on normal images cannot be generalized to abnormal images, in other words, i.e. it cannot reconstruct the abnormal image as well as the normal ones.

The deep generative models based on variational autoencoder (VAE)~\cite{VAE}~and generative adversarial network (GAN)~\cite{gan}~can also be used as the reconstruction model. Baur et al.~\cite{vae-l1} propose VAE-GAN which is used to detect suspicious lesions in brain magnetic resonance images (MRI). There, GAN is used for adversarial training to improve the quality of the reconstructed image. During testing, VAE-GAN only uses pixel-level~$l_1$-distance to score the abnormality. Schlegl et al.~\cite{f-anogan} adopt a similar method. They take convolution autoencoder instead of variational autoencoder and detect anomalies for optical coherence tomography image.
In addition to using the common distance to measure the pixel difference, anomaly detection methods based on the deep generative model can further leverage the reconstruction probability~\cite{ae-ssim-l2,vae-prob} or likelihood score~\cite{gan-avid,gan-prl}~as additional anomaly measures.

Besides, some methods detect anomalies not by comparing the differences between the input image and the reconstructed image, but by calculating the differences between the input image and its nearest normal image.
Schlegl et al.~\cite{anogan} propose AnoGAN. Firstly, a GAN model is trained with normal images. Then the anomaly is detected by estimating the difference between the test image and the normal image which is closest to the test image. This normal image is obtained through an iterative optimization process. Firstly, the latent code that is closest to the test image's is searched in the latent space of the GAN model. Then the corresponding normal image is generated with the generator of the GAN with the obtained latent code.
David et al.~\cite{vae-grad} also propose to detect anomalies by comparing the difference between the test image and its nearest neighbor normal image obtained by an optimization. 
Since both methods need to perform an iterative search process, they are inefficient for practical applications.
%Then, by optimizing a loss function defined on the test image and the output of the variable self encoder, the gradient descent strategy is used to update the input of the variable self encoder step by step until the closest image is found for the test image, that is, the input of the variable self encoder at the end of the iteration.

Moreover, some methods propose to increase the difficulty of reconstruction~\cite{yang2020learning,zavrtanik2021reconstruction}. They remove some information of the input image firstly, and then make the convolutional autoencoder to reconstruct the input image with the incomplete or degraded input. The information degradation process can effectively increase the difficulty of abnormal image reconstruction, so it can enlarge the anomaly score between the normal and abnormal samples, thus improving the detection performance.

Image reconstruction for pixel-level anomaly detection is very intuitive. But because it detects potential anomalies by evaluation the pixel-wise differences in the pixel or image space, it usually is expected to regenerate high-quality images for comparisons. However, high quality image generation itself is still a challenging task.
For example, the reconstruction approaches usually struggle to regenerate the sharp edges and complex texture structure for images. As a result, large reconstruction errors often appear in the edge or texture regions, leading to a large number of false abnormal alarms.

\subsection{Feature modeling}
In stead of detecting anomaly in the image space, the feature-based methods detect anomaly in the feature space. This category of methods is devoted to constructing the effective representation of the local regions of the whole image by using the handcrafted features~\cite{texem-gmm,texem-gmm-cs,atom-fd,multi-atom-fd}~or the representation learned by neural networks~\cite{cs-fd,cnn-fd,ST,dfr}.
Then, many machine learning models such as sparse coding~\cite{atom-fd,multi-atom-fd,cs-fd}, Gaussian mixed model~\cite{texem-gmm,texem-gmm-cs}and Kmeans clustering~\cite{cnn-fd} can be used to model the feature distribution of normal images.
For anomaly detection, if the regional feature corresponding to the local region of the test image deviate from the modeled feature distribution, this region will be labeled as abnormal.

In order to improve the detection performance, the multi-scale model ensemble strategy~\cite{texem-gmm,multi-atom-fd} is usually adopted which combines the results of multiple single models derived from different image region sizes.
Besides, to locate the abnormal region in the image, feature-based methods usually needs to divide the image into many small image patches firstly, and then model and detect the anomalies in image-patch level. Therefore, it is very time-consuming during both training and testing, especially when the deep neural network is required to extract the deep image features.
Mover, because each local region of an image is evaluated independently, it may not be able to infer anomalies by leveraging the global context information of the image. Although the proposed multi-scale modeling strategy can alleviate the mentioned shortcomings~\cite{texem-gmm,multi-atom-fd}. However, the multi-scale strategy implicitly assumes that each scale is independent and totally ignores the relationship between different feature scales which is very important for making a comprehensive detection decision.

Recently, Bergmann et al.~\cite{mvtec} developed a benchmark dataset for evaluating unsupervised image anomaly detection algorithms: MVtec AD. The dataset consists of various texture and object categories, and contains more than 70 different types of anomalies. The authors evaluated many methods based on image reconstruction and feature modeling on the proposed dataset, and found that the detection performances of these methods on different data categories are quite unstable, which shows an considerable room for improvement.
Later, Bergmann et al.~\cite{ST} further proposed an unsupervised anomaly detection method with a student-teacher distillation framework, and achieved very promising results on MVtec AD dataset.
They leverage the transferred deep convolution features and detect the anomaly through feature regression.
Specifically, they take the pre-trained deep convolution features (such as ResNet18~\cite{resnet}) as the regression target or the teacher, and train a group of student networks that imitate the teacher's behavior or regress the target features of the normal images.
During testing, it first uses the student networks to predict the teacher's network output, and then calculates the anomaly score according to the corresponding prediction errors and uncertainties (variance).
This method assumes that the student network only trained to regress the output of the teacher network on the normal images may not be able to predict the output of the teacher network on the abnormal images.
%In addition, the author also suggests that multi-scale modeling strategy be used to further improve the anomaly detection performance of the model.

Recently, some feature-based methods, such as~\cite{dfr,patchsvdd,spade}, are devoted to coding multiple spatial context information by making use of pre-trained deep hierarchical convolution features, which have shown great potential for pixel-level anomaly detection and segmentation.
Shi et al.~\cite{dfr} propose to make full use of the hierarchical features of the pre-trained deep convolution network, such as VGG19~\cite{vgg19} on Imagenet~\cite{imagenet}, and develop an effective feature reconstruction mechanism for anomaly detection.
Yi et al.~\cite{patchsvdd}~propose a hierarchical convolutional encoder which can extract hierarchical features and designs a self-supervised training strategy. For anomaly detection, they specially develop a feature matching mechanism. Concretely, the features of normal image patches are extracted and stored beforehand. Then, in the reference stage, the features of the test image will be matching with the features those have been stored previously by the nearest neighbor searching method. Finally, the minimum matching distance is taken as the anomaly score of each patch of the test image.
Cohen et al.~\cite{spade} adopt a similar feature matching mechanism. Instead, they suggest using the pre-trained deep pyramid convolution features, such as the convolution feature maps from ResNet18~\cite{resnet} on Imagenet~\cite{imagenet}.

Another research line of visual anomaly detection for images is leveraging the gradient-based attention mechanism, such as Grad-CAM~\cite{grad-cam}and interpretable deep generative models~\cite{evae,cavga}. Venkataramanan et al.~\cite{cavga} develop an guided attention mechanism to locate the abnormal regions in the image. Liu et al.~\cite{evae} propose a gradient-based visual interpretation approach to estimate the potential anomalies in the image with the variational autoencoder that only trained on the normal data.

\section{Conclusion}
In this survey paper, we have discussed various visual anomaly detection approaches. The goal of this survey is to investigate and identify the underlying principles and assumptions of the different models for anomaly detection and help the researchers to understand the current research status and trends of visual anomaly detection, draw insights from the current approaches and open up new research directions.


% % use section* for acknowledgment
% \section*{Acknowledgment}

% This work is supported by grants from: National Natural Science Foundation of China (No.71932008, 91546201 and 71331005).



\bibliographystyle{IEEEtran}
\bibliography{bare_conf}


% \end{thebibliography}

\end{document}